\documentclass[runningheads]{llncs}
\usepackage{amsfonts}
\usepackage{amsmath}
\usepackage{graphicx}
\usepackage{nicefrac}
\usepackage{multirow}
\begin{document}
%
\title{Understanding  the classes better with class-specific and rule-specific feature selection, and redundancy control in a fuzzy rule based framework}
\titlerunning{ }
%
\author{Suchismita Das \orcidID{0000-0002-3952-3605} \and
		Nikhil R. Pal\orcidID{0000-0001-6935-901X}}
\authorrunning{S. Das and N. R. Pal}
%
\institute{Electronics and Communication Sciences Unit\\Indian Statistical Institute, Calcutta, 700108, India \\
		\email{\{suchismitasimply,nrpal59\}@gmail.com}\\}

\maketitle              
\begin{abstract}
Recently, several studies have claimed that using class-specific feature subsets provides certain advantages over using a single feature subset for representing the data for a classification problem. Unlike traditional feature selection methods, the class-specific feature selection methods select an optimal feature subset for each class. Typically  class-specific feature selection  (CSFS) methods use one-versus-all split of the data set that leads to issues such as class imbalance, decision aggregation, and high computational overhead. We propose a class-specific feature selection method embedded in a fuzzy rule-based classifier, which  is free from the  drawbacks associated with most  existing class-specific  methods. Additionally, our method can be adapted to control the level of redundancy in the class-specific feature subsets by adding a suitable regularizer to the learning objective. Our method results in class-specific rules involving class-specific subsets. We also propose an extension where different rules of a particular class are defined by different feature subsets to model different substructures within the class. The effectiveness of the proposed   method has been validated through experiments on three synthetic data sets.  

\end{abstract}

\section{Introduction}
Feature selection is an important  step for many  machine learning tasks. The main motto of the feature selection methods is to reject unnecessary and derogatory features and select the features that benefit the intended task. 
Traditional feature selection methods choose a single subset of the  features as the ``optimal" subset for the entire dataset. Apart from the commonly used global approach, some studies \cite{de2011class,ezenkwu2021class,nardone2019sparse,panthong2019liver,pineda2011general,qian2019class,yuan2020using,zhou2014novel} have used class-specific approaches for selecting features, where for each class, a unique subset of the original features is selected. If there are $C$ classes, in the class-specific approach, $C$ subsets are chosen. In a traditional feature selection method, the selected feature subset is chosen based on the global characteristics of the data. It does not take into account  any class-specific or local characteristics of the data which may be present. For example, there may exist a group of features that follows a distinct distribution for a specific class but varies randomly over the remaining classes. Such a group of features plays a significant role in distinguishing the specific class from other classes. However, such a group  may not be a very useful feature subset for the $C$-class problem as a whole. Class-specific characteristics may also exist in the form of class-specific redundancy. Different sets of features could be redundant for different classes. 
The class-specific feature selection (CSFS)  works in  \cite{de2011class,ezenkwu2021class,nardone2019sparse,panthong2019liver,pineda2011general,qian2019class,yuan2020using,zhou2014novel}  have proposed suitable frameworks that exploit class-specific feature subsets to solve  classification problems. They have shown that the classifiers built with the subsets chosen by  class-specific  methods performed better than or comparable to the classifiers built with subsets chosen by the traditional global feature selection methods. The CSFS may also enhance the transparency/explainability of the classification process associated with it \cite{ezenkwu2021class}. 

The majority of the CSFS methods \cite{de2011class,ezenkwu2021class,panthong2019liver,pineda2011general,yuan2020using,zhou2014novel} follow the one-versus-all (OVA) strategy to decompose a $C$-class classification problem into $C$ binary classification problems. They choose $C$ class-specific feature subsets optimal for the $C$ binary classification problems.  OVA strategy-based class-specific feature selection methods have certain drawbacks. Generally, it leads to class imbalance.  The performance of an OVA-based method would depend on how efficiently the  class imbalance problem is handled. The   OVA strategy-based methods is computationally intensive
and complex as we need to design $C$ classifiers and for testing we need an aggregation mechanism.
 
Here, we propose a CSFS scheme embedded in a fuzzy rule-based classifier (FRBC) that does not use the OVA strategy. Our method selects class-specific feature subsets by learning a single FRBC and hence avoid the issues introduced in the OVA-based approaches. Moreover, we extend our  framework to deal with- (i) CSFC  with redundancy control, and (ii) rule-specific feature selection (RSFS) that can exploit  presence of substructure within a class.  The rules provided by the FRBC are generally interpretable and more specific. Exploiting  the class-specific local features, the proposed FRBC enjoys more transparency and interpretability than a standard classifier exploiting class-specific feature subsets. Our contributions are summarised  as follows.  
\begin{enumerate}
	\item We propose a class-specific feature selection method that is not based on the OVA strategy like most of the existing class-specific feature selection schemes. Thus, our method is free from the weaknesses of the OVA strategy.
	\item Our method can monitor the level of redundancy in the selected features.
	\item We also propose a general version of our class-specific feature selection method that not only chooses different subsets for different classes but also, chooses different subsets within a class if different substructures are present.
\end{enumerate}                      
\section{Proposed Method}\label{sec:prop}
We want to develop a method for CSFS based on a training dataset. Let, the input data be $\mathbf{X}=\{\mathbf{x}^{i}=(x_{1}^{i},x_{2}^{i},\cdots,x_{P}^{i})^{T}\in \mathcal{R}^{P}: i \in \{1,2,\cdots,n\}\}$. 
The collection of class labels of  $\mathbf{X}$ be $\mathbf{y}=\{y^{i} \in\{1,2,\cdots, C\}:i \in \{1,2,\cdots,n\}\}$, where, $y^{i}$ is the class label corresponding to $\mathbf{x}^{i}$. For our purposes we represent the class label of $\mathbf{x}^{i}$ as $\mathbf{t}^{i} \in \{0,1\}^{C}$, where ${t}_{k}^{i}=1$ if $y^{i}=k$ and ${t}_{k}^{i}=0$, otherwise. 
We denote the $j$th feature by $x_{j}$, the class label  by $y$, and the target vector by $\mathbf{t}$. The set of original features be $\mathbf{f}=\{x_{1}, x_{2}, \cdots x_{P}\}$. Let the optimal class-specific subset for the $k$th class be $\mathbf{s}_k$. We need to find out $\mathbf{s}_{k}$s where, $\mathbf{s}_{k} \subset \mathbf{f} \hspace{2pt}\forall k \in\{1,2,\cdots, C\}$. We propose a CSFS mechanism embedded in a fuzzy rule-based classifier (FRBC). So next we discuss the FRBC.
\subsection{FRBC}\label{subsec:FRBC}
We  employ the FRBC framework used in \cite{chen2011integrated,chung2017feature}. Each class is represented by a set of rules. Let there are $N_{k}$ rules for the $k$th class. The $l$th rule corresponding to the $k$th class, $\mathrm{R}_{kl}$ is given by
\begin{equation}
	\mathrm{R}_{kl}:\texttt{If } x_{1} \texttt{ is } A_{1,kl} \texttt{ and } x_{2} \texttt{ is } A_{2,kl} \texttt{ and} \cdots x_{P} \texttt{ is } A_{P,kl} \texttt{then } y \texttt{ is } k. \label{eq:rule} 
\end{equation} 
Here, $k\in\{1,2,\cdots, C\}$, $l \in \{1,2,\cdots, N_{k}\}$, and $A_{j,kl}$ is a linguistic value (fuzzy set) defined on the $j$th feature for the $l$th rule of the $k$th class. Let, $\alpha_{kl}$ be the firing strength of the rule $\mathrm{R}_{kl}$. The rule firing strength is computed using a $T$-norm \cite{gupta1991theory} over the fuzzy sets $A_{1,kl},A_{2,kl},\cdots A_{P,kl}$. We use the product  $T$-norm. Let the membership to the fuzzy set $A_{j,kl}$ be $\mu_{j,kl}$. So, $\alpha_{kl}$ is given by, $\alpha_{kl}=\prod_{j=1}^{P} \mu_{j,kl}$. The final output of the FRBC is of the form $\mathbf{o}=(o_{1}, o_{2}, \cdots, o_{C})$, where, $o_{k}$ is the support for $k$th class, computed as $o_{k}=\max \{\alpha_{k1}, \alpha_{k2}, \cdots, \alpha_{kN_{k}}\}$. To learn an efficient classifier from the initial fuzzy rule-based system, the  parameters defining the fuzzy sets $A_{j,kl}$s can be tuned by minimizing the classification error, 
\begin{equation}
E_{cl}=\textstyle  {\sum_{i=1}^{n}\sum_{k=1}^{C}(o^{i}_{k}-t^{i}_{k})^2}. \label{eq:error_cl}
\end{equation}    

To extract the $l$th rule of the $k$th class, $\mathrm{R}_{kl}$ we need to define the fuzzy sets $A_{j,kl}$s. Following \cite{chen2011integrated,chung2017feature}, we cluster the training data of the $k$th class into $N_{k}$ clusters. We  note here that the $k$th class  may not have $n_{k}$  clusters in the pattern recognition sense. By clustering we  just group the nearby points  and then define a rule for each group. Let the centroid of the  $l$th cluster  of the $k$th class be represented as $\mathbf{v}_{kl}=(v_{1,kl}, v_{2,kl},\cdots,v_{P,kl})$. The cluster centroid $\mathbf{v}_{kl}$ is then translated into $P$ fuzzy sets, $A_{j,kl}=\text{ ``close to" } v_{j,kl} \hspace{3pt} \forall j \in \{1,2,\cdots,P\} $.  The fuzzy set `$\text{ ``close to" } v_{j,kl}$' is modeled by a Gaussian membership function with mean $v_{j,kl}$. Although the membership parameters  can be tuned to refine the fuzzy rules, in this study, we have not done that. We have used fixed rules defined by the obtained cluster centers. 
\subsection{Feature Selection}
Following \cite{banerjee2015unsupervised,chakraborty2014feature,chen2011integrated,chung2017feature}, we use feature modulators which stop the derogatory features and promote useful features to take part in the rules of the FRBC. We choose the modulator function same as used in \cite{chung2017feature}. For each feature, there is an associated modulator of the form $M(\lambda_{j})=\exp{(-\lambda_{j}^{2})}$, where $j \in \{1,2,\cdots,P\}$.  To select or reject a feature using the modulator function, the membership values associated with the $j$th feature are modified as
\begin{equation}
	\textstyle{\hat{\mu}_{j,kl}=\mu_{j,kl}^{M(\lambda_{j})}=\mu_{j,kl}^{\exp{(-\lambda_{j}^{2})}} \forall k,l} \label{eq:mem}
\end{equation}
Note that, $\lambda_{j} \approx 0$ makes $\hat{\mu}_{j,kl}\approx\mu_{j,kl}$. Similarly, when $\lambda_{j}$ is high (say, $\lambda_{j} \geq 2$), $\hat{\mu}_{j,kl}\approx 1$. The rule firing strength is now calculated as $\alpha_{kl}=\prod_{j=1}^{P} \hat{\mu}_{j,kl}$. So, when $\hat{\mu}_{j,kl}\approx\mu_{j,kl}$, $j$th feature  influences  the rule firing strength $\alpha_{kl}$ and in turn influences the classification process, whereas, if $\hat{\mu}_{j,kl}\approx 1$ then the $j$th feature has no influence on the firing strength and hence on the predictions by the FRBC. This would be true for any T-Norm as $T(x,1)=x, x \in [0,1]$. Thus, for useful features, $\lambda_{j}$s should be made close to zero  and for derogatory features $\lambda_{j}$s should be made high. The desirable values of $\lambda_{j}$s are obtained by minimizing $E_{cl}$ defined in (\ref{eq:error_cl}) with respect to $\lambda_{j}$s. The training begins with $\lambda_{j}=2+$ Gaussian noise. $M(\lambda_{j}) \approx 0$ indicates a strong rejection of $x_j$, while $M(\lambda_{j}) \approx 1$  suggests a strong acceptance of $x_j$. However, training may lead $\lambda_{j}$s such that  $M(\lambda_{j})$ takes a value in between $0$ and $1$. This implies that the corresponding feature influences the classification partially. This is not desirable in our case, as our primary goal is to select or reject features.     To facilitate this, we add a regularizer term $E_{select}$ to $E_{cl}$ such that $E_{select}$ adds penalty if any $\lambda_{j}$ allows the corresponding feature partially. In \cite{chung2017feature}, $E_{select}$ is set as follows.
\begin{equation}
	E_{select}=\textstyle  {(\nicefrac{1}{P})\sum_{j=1}^{P}\exp{(-\lambda_{j}^{2})}(1-\exp{(-\lambda_{j}^{2})})} \label{eq:sel}
\end{equation} 
So, the overall loss function for learning suitable $\lambda_{j}$s  becomes
\begin{equation}
	\textstyle  {E= E_{cl}+ c_{1}E_{select}} \label{eq:error}.
\end{equation} 
\subsubsection{Class-specific feature selection} 
So far we have not considered selection of class-specific features.
In the class-specific scenario, for each class, a different set of $P$  modulators is engaged. So, a total of $C\times P$ feature modulators are employed. Consequently, for each class a different set of features, if appropriate, can be selected. 
Here, we represent the feature modulator for the $j$th feature of the $k$th class as $M(\lambda_{j,k})=\exp{(-\lambda_{j,k}^{2})}$, where $j \in \{1,2,\cdots,P\}; k \in \{1,2,\cdots,C\}$. The modulator value $M(\lambda_{j,k})$ modify the membership values corresponding to the $j$th feature of the $k$th class as following:
\begin{equation}
	\textstyle  {\hat{\mu}_{j,kl}=\mu_{j,kl}^{M(\lambda_{j,k})}=\mu_{j,kl}^{\exp{(-\lambda_{j,k}^{2})}} \forall l} \label{eq:mem_cs} 
\end{equation} 
For this problem, the $E_{select}$ is changed to
\begin{equation}
	E_{select}=\textstyle  {(\nicefrac{1}{CP})\sum_{k=1}^{C}\sum_{j=1}^{P}\exp{(-\lambda_{j,k}^{2})}(1-\exp{(-\lambda_{j,k}^{2})})} \label{eq:sel_cs}
\end{equation}       
We now Minimize (\ref{eq:error}) with respect to $\lambda_{j,k}$s to find  the optimal $\lambda_{j,k}$s. 
\subsection{Monitoring Redundancy}
Suppose a data set has three useful features say $x_{1},x_{2},x_{3}$ such that each of $x_2$ and $x_3$ is  strongly dependent on (say correlated with)  $x_1$ then all the three features carry the same information and only one of them is enough. These three form a redundant set of features. However, if we just use one of them and there is some error in measuring that feature, the system may fail to do the desired job. Therefore, instead of minimizing redundancy, a controlled use of redundant features is desirable.
 For the global feature selection framework, redundancy control has been realized using the feature modulators      \cite{banerjee2015unsupervised,chakraborty2014feature,chung2017feature} by adding the  regularizer \eqref{eq:red}  to (\ref{eq:error}):  
\begin{equation}
	E_{r}=\textstyle  {(\nicefrac{1}{P(P-1)})\sum_{j=1}^{P}\sum_{m=1, m\neq j}^{P} \sqrt{\exp{(-\lambda_{j}^{2})}\exp{(-\lambda_{m}^{2})}\rho^{2}(x_{j},x_{m})}} \label{eq:red}
\end{equation}
Here, $\rho()$ is the Pearson’s correlation coefficient, which is a measure of dependency between two features. When $x_{j}$ and $x_{m}$ are highly correlated, $\rho^{2}(x_{j},x_{m})$ is close to one (its highest value). In this case, to reduce the penalty $E_{r}$, the training process will adapt  $\lambda_{j}$ and $\lambda_m$ in such a way  that    one of  $\exp{(-\lambda_{j}^{2})}$ and $\exp{(-\lambda_{m}^{2})}$ is close to $0$ and the other is close to $1$. Note the    (\ref{eq:red}) is not suitable for class-specific scenario. Next we change (\ref{eq:red}) for class-specific redundancy.
\subsubsection{Class-specific redundancy}
For class-specific redundancy, we need to compute class-specific dependency of a feature pair. So we compute $\rho_{k}(x_{j},x_{m})$ between   features $x_{j}$ and $x_{m}$ considering only instances of the $k$th class. In the class-specific case, for each class, we have $P$ feature modulators, $M(\lambda_{j,k})=\exp{(-\lambda_{j,k}^{2})}$, where $j \in \{1,2,\cdots,P\}; k \in \{1,2,\cdots,C\}$. So, (\ref{eq:red}) is modified as following.
\begin{equation}
	E_{r_{c}}=\textstyle  {(\nicefrac{1}{CP(P-1)})\sum_{k=1}^{C}\sum_{j=1}^{P}\sum_{m=1, m\neq j}^{P} \sqrt{\exp{(-\lambda_{j,k}^{2})}\exp{(-\lambda_{m,k}^{2})}\rho_{k}^{2}(x_{j},x_{m})}} \label{eq:red_cs}
\end{equation} 
Considering the class-specific redundancy, our new loss function for learning the system becomes:
\begin{equation}
	\textstyle  {E_{tot}=E_{cl}+c_{1}E_{select}+c_{2}E_{r_c}}. \label{eq:error_tot}
\end{equation}       
\subsection{Exploiting Substructures Within a Class.}
For some real world problems, the data corresponding to a class may have distinct clusters and some of the clusters may lie in different sub-spaces. For example,in a multi-cancer gene expression data set, each cancer may have several sub-types, where each sub-type is characterized by a different set of genes/features. This generalizes the concept of class-specific feature selection further. To exploit such local substructures within a class while extracting rules, we need to use rule-specific feature modulators. 
Each rule of the $k$th class is assumed to represent a local structure or cluster present in the $k$th class.  
So, for the $k$th class there are $n_{k}\times P$ feature modulators. For the overall system there are $n_{rule}\times P$ feature modulators where, $n_{rule}(=\sum_{k=1}^{C}n_{k})$ is the total  number of rules. A modulator function is now represented by $M(\lambda_{j,kl})$ and the corresponding modulated membership is the following. 
\begin{equation}
	\textstyle  {\hat{\mu}_{j,kl}=\mu_{j,kl}^{M(\lambda_{j,kl})}=\mu_{j,kl}^{\exp{(-\lambda_{j,kl}^{2})}}} \label{eq:mem_rule}
\end{equation} 
The regularizer, $E_{select}$ is now modified as
\begin{equation}
	E_{select}=\textstyle  {(\nicefrac{1}{CP})\sum_{k=1}^{C}(\nicefrac{1}{n_{k}})\sum_{l=1}^{n_{k}}\sum_{j=1}^{P}\exp{(-\lambda_{j,kl}^{2})}(1-\exp{(-\lambda_{j,kl}^{2})})}\label{eq:sel_rs}
\end{equation}
In this framework,  we do not consider redundancy. Using (\ref{eq:mem_rule}) for  $E_{cl}$ and (\ref{eq:sel_rs}) for $E_{select}$ we define the loss function $E= E_{cl}+ c_{1}E_{select}$ for discovering rule-specific feature subset.  
\section{Experimentation}
We do three experiments to validate three main contributions of our proposed framework. In Experiment 1, we  show effectiveness of the proposed class-specific feature selection over the usual global feature selection using a FRBC. In Experiment 2, we  demonstrate the significance of class-specific redundancy control using our approach. In Experiment 3, a data set having multiple sub-structures in different sub-spaces within a class is considered to show the utility of our method.  We do not tune the rule base parameters of the FRBC and only tune the feature modulators to select/reject features. For clustering, we  use the $K$-means  algorithm. To minimize the error functions using stochastic gradient descent, we  use the optimizer, \texttt{train.GradientDescentOptimizer} from    \texttt{TensorFlow} \cite{abadi2016TensorFlow}. For all experiments, the learning rate  is set to $0.2$. As mentioned in sec.\ref{sec:prop} we denote the class-specific feature subset for class 1 as $\mathbf{s}_{1}$, for class 2 as $\mathbf{s}_{2}$ and so on.
\subsection{Experiment 1}
For the first experiment, we have considered a three class synthetic data set Synthetic1 with six features having distributions as described in Table \ref{tab:tab1}.    
\begin{table}[h!]
	\caption{Description of the dataset Synthetic1}
	\label{tab:tab1}
		\centering
		\begin{tabular}{|c| c c c c c c |c|}
			\hline
			\multirow{2}{*}{Instances} & \multicolumn{6}{c|}{Features} &  Class\\
			&  $x_1$ & $x_2$ & $x_3$ & $x_4$ & $x_5$ & $x_6$ & $\mathbf{y}$\\ 
			\hline
			
			\rule{0pt}{3ex}$\mathbf{x}^1 \cdots \mathbf{x}^{100}$ & \scalebox{0.9}{ $\mathcal{N}(0,0.5)$} & \scalebox{0.9}{$\mathcal{N}(0,0.5)$} & \scalebox{0.9}{$\mathcal{U}(-10,10)$} & \scalebox{0.9}{$\mathcal{U}(-10,10)$} &\scalebox{0.9}{$\mathcal{U}(-10,10)$} & \scalebox{0.9}{$\mathcal{U}(-10,10)$} & $1$  \\
			\rule{0pt}{3ex}$\mathbf{x}^{101} \cdots \mathbf{x}^{200}$ & \scalebox{0.9}{$\mathcal{U}(-10,10)$} & \scalebox{0.9}{$\mathcal{U}(-10,10)$} & \scalebox{0.9}{$\mathcal{N}(0,0.5)$} & \scalebox{0.9}{$\mathcal{N}(0,0.5)$} & \scalebox{0.9}{$\mathcal{U}(-10,10)$} &\scalebox{0.9}{$\mathcal{U}(-10,10)$} & $2$  \\
			\rule{0pt}{3ex}$\mathbf{x}^{201} \cdots \mathbf{x}^{300}$ & \scalebox{0.9}{$\mathcal{U}(-10,10)$} & \scalebox{0.9}{$\mathcal{U}(-10,10)$} & \scalebox{0.9}{$\mathcal{U}(-10,10)$} & \scalebox{0.9}{$\mathcal{U}(-10,10)$} & \scalebox{0.9}{$\mathcal{N}(0,0.5)$} & \scalebox{0.9}{$\mathcal{N}(0,0.5)$} & $3$  \\
			\hline
		\end{tabular}
\end{table}
Here, $\mathcal{N}(m,s)$ represents a normal distribution with mean,  $m$ and standard deviation,  $s$; $\mathcal{U}(a,b)$ represents a uniform distribution over the interval $(a,b)$. Without  loss, we have assigned the first $100$ points to class $1$, next $100$ points to class 2 , and last $100$ points to class 3. From Table \ref{tab:tab1} we can see that class 2 and class 3 are uniformly distributed over a given interval for features $x_{1}$ and $x_{2}$. On the other hand, class 1 is clustered around $(0,0)$ in the feature space formed by $x_{1}$ and $x_{2}$. Hence the feature space formed by $x_{1}$ and $x_{2}$, discriminate class 1 from the other two classes. Similarly, the feature spaces formed of $(x_3,x_4)$ and $(x_{5},x_{6})$ discriminate class 2 and class 3 respectively, from the corresponding remaining classes. To understand the importance of CSFS, we perform both global feature selection (GFS) and CSFS,  and compare their performances. We have also computed the performance of the FRBC with all features. Number of rules considered per class is one. We have conducted $5$ runs for each of the FRBC. 
\begin{table}[h!]
	\caption{Performance on Synthetic1}
	\label{tab:tab2}
	\centering
	\begin{tabular}{|c|c| c| c |c| c|}
		\hline
		 \multirow{2}{*}{Run}&\multicolumn{2}{c|}{Features selected} & \multicolumn{3}{c|}{ Avg. accuracy of FRBC (\%)} \\\cline{2-6}
		 
		  &Class-specific & Global & Class-specific & Global &  All features\\
		\hline
	     1-5 & $\mathbf{s}_1:x_{1},x_{2}$; $\mathbf{s}_{2}:x_{3},x_{4}$; $\mathbf{s}_{3}:x_{5},x_{6}$ & $x_{3},x_{6}$ &98.7 &34.7 &62.08\\ 
		\hline
	\end{tabular}
\end{table} 
  We observe from Table \ref{tab:tab2} that in class-specific feature selection, in all five runs, for each class its characteristic features (i.e. $x_{1}, x_{2}$ for class 1 and so on) are selected. The FRBC with the class-specific selected features  has achieved an average accuracy of $98.7\%$ - in fact, each run achieved the same accuracy. Whereas, in global feature selection, the selected subset is $x_{3},x_{6}$. The FRBC using globally selected feature subset has achieved an accuracy of  of $34.7\%$ in each of the five runs. One can argue  that the class-specific model uses all  six features, hence performs better than the global model which uses two features. But, when we learn the FRBC rules using all  six  features it has achieved an  average accuracy of $62.08\%$ over the five runs. Importance of class-specific feature selection is clearly established through this experiment. 
\subsection{Experiment 2}
For Experiment 2, we have considered another synthetic dataset Synthetic2 which is produced by appending two additional features $x_{7}$ and $x_{8}$ to  Sythetic1 data set. For class 1, $x_{7}$ and $x_{8}$ are generated as $x_{1}+\mathcal{N}(0,0.1)$ and $x_{2}+\mathcal{N}(0,0.1)$, respectively.  For the other two classes, $x_{7}$ and $x_{8}$  are generated from $\mathcal{U}(-10,10)$ and $\mathcal{U}(-10,10)$, respectively. We observe that  $x_{7}$ is dependent on $x_{1}$ and  $x_{8}$ is dependent on $x_{2}$ for class 1 but the remaining two classes are indiscernible among themselves considering a feature space with $x_{7}$ and $x_{8}$. Clearly, the features $x_{7},x_{8}$ are also discriminatory for class 1. However, do $x_{7},x_{8}$ add any information over $x_{1},x_2$ for class 1? The answer is no, as for class 1, $x_{7}$ and $x_{8}$ are noisy versions of $x_{1}$ and $x_{2}$, respectively. This feature redundancy  is specific to class 1. In Table \ref{tab:tab4} we have described the performances of the  FRBCs in the CSFS framework without and with class-specific redundancy control. 
\begin{table}[h!]
	\caption{Class-specific feature selection with  and without class-specific redundancy control on Synthetic2 data set}
	\label{tab:tab4}
	\centering
	\begin{tabular}{|c|c|c|c|c|}
		\hline
		\multirow{2}{*}{Run} & \multicolumn{2}{c|}{Without redundancy control} & \multicolumn{2}{c|}{With class-specific redundancy control}\\ \cline{2-5}
		 & Selected features & Acc.& Selected features & Acc.\\
		\hline
		1& $\mathbf{s}_{1}$:$x_{1},x_{2},x_{7},x_{8}$; $\mathbf{s}_{2}$:$x_{3},x_{4}$; $\mathbf{s}_{3}$:$x_{5},x_{6}$ & 99.3&$\mathbf{s}_{1}$:$x_{2},x_{7}$; $\mathbf{s}_{2}$:$x_{3},x_{4}$; $\mathbf{s}_{3}$:$x_{5},x_{6}$ & 99.3 \\ 
		2& $\mathbf{s}_{1}$:$x_{1},x_{2},x_{7}$; $\mathbf{s}_{2}$:$x_{3},x_{4}$; $\mathbf{s}_{3}$:$x_{5},x_{6}$ & 99.3&$\mathbf{s}_{1}$:$x_{1},x_{8}$; $\mathbf{s}_{2}$:$x_{3},x_{4}$; $\mathbf{s}_{3}$:$x_{5},x_{6}$ & 99 \\ 
		3&$\mathbf{s}_{1}$:$x_{1},x_{2},x_{7},x_{8}$; $\mathbf{s}_{2}$:$x_{3},x_{4}$; $\mathbf{s}_{3}$:$x_{5},x_{6}$ & 99.3&$\mathbf{s}_{1}$:$x_{7},x_{8}$; $\mathbf{s}_{2}$:$x_{3},x_{4}$; $\mathbf{s}_{3}$:$x_{5},x_{6}$ & 99.3 \\ 
		4&$\mathbf{s}_{1}$:$x_{1},x_{2},x_{7},x_{8}$; $\mathbf{s}_{2}$:$x_{3},x_{4}$; $\mathbf{s}_{3}$:$x_{5},x_{6}$ & 99.3&$\mathbf{s}_{1}$:$x_{7},x_{8}$; $\mathbf{s}_{2}$:$x_{3},x_{4}$; $\mathbf{s}_{3}$:$x_{5},x_{6}$ & 99.3 \\
		5&$\mathbf{s}_{1}$:$x_{1},x_{2},x_{7},x_{8}$; $\mathbf{s}_{2}$:$x_{3},x_{4}$; $\mathbf{s}_{3}$:$x_{5},x_{6}$ & 99.3&$\mathbf{s}_{1}$:$x_{1},x_{7}$; $\mathbf{s}_{2}$:$x_{3},x_{4}$; $\mathbf{s}_{3}$:$x_{5},x_{6}$ & 99 \\ 
		\hline
	\end{tabular}
\end{table}
Here also, we have set the number of fuzzy rules per class as one and repeated the experiments five times with each model. The term `Acc.' mentioned in Table \ref{tab:tab4} refers to accuracy of the FRBC in percentage. Table \ref{tab:tab4} confirms the effectiveness of using class-specific redundancy control. For class 1, features $x_{1}$ and $x_{7}$ are heavily dependent. Hence, to avoid redundancy only one of them should be selected. The same argument is true for features $x_2$ and $x_8$. Using an objective function (\ref{eq:error_tot}) which considers a regularizer on class-specific redundancy associated penalty (\ref{eq:red_cs}), the FRBC has successfully chosen only one from $x_{1}$ and $x_{7}$ and one from $x_2$ and $x_8$ to include in $\mathbf{s}_{1}$ in all five runs.     
On the other hand, we observe that without any redundancy control, the class-specific feature selection framework selects all the four discriminating features to include in $\mathbf{s}_{1}$ in four runs. The best accuracy achieved by the CSFS framework without any redundancy control and that of CSFS  with class-specific redundancy control are same and equal to $99.3\%$ although the later selects only two features. This experiment establishes the benefit of class-specific redundancy control.
%
\subsection{Experiment 3}
In experiment 3, we validate our proposed framework for handling the presence of different clusters or structures in different sub-spaces within a class. We have synthesized, Synthetic3, a two class data having four features where each class is composed of two distinct clusters lying in two different sub-spaces. The data set Synthetic3 is described in Table \ref{tab:tab6}.
\begin{table}[h!]
	\caption{Description of the dataset Synthetic3}
	\label{tab:tab6}
	\centering
	\begin{tabular}{|c| c c c c  |c|}
		\hline
		Instances & \multicolumn{4}{c|}{Features} &  Class\\
		&  $x_1$ & $x_2$ & $x_3$ & $x_4$ &  $\mathbf{y}$\\ 
		\hline
		
		\rule{0pt}{3ex}$\mathbf{x}^1 \cdots \mathbf{x}^{100}$ & \scalebox{0.7}{ $\mathcal{N}(0,0.5)$} & \scalebox{0.7}{$\mathcal{N}(-5,0.5)$} & \scalebox{0.7}{$\mathcal{U}(-10,10)$} & \scalebox{0.7}{$\mathcal{U}(-10,10)$} & $1$  \\
		\rule{0pt}{3ex}$\mathbf{x}^{101} \cdots \mathbf{x}^{200}$ & \scalebox{0.7}{$\mathcal{U}(-10,10)$} & \scalebox{0.7}{$\mathcal{U}(-10,10)$} & \scalebox{0.7}{$\mathcal{N}(0,0.5)$} & \scalebox{0.7}{$\mathcal{N}(0,0.5)$} &  $1$  \\
		\rule{0pt}{3ex}$\mathbf{x}^{201} \cdots \mathbf{x}^{300}$ & \scalebox{0.7}{$\mathcal{U}(-10,10)$} & \scalebox{0.7}{$\mathcal{N}(0,0.5)$} & \scalebox{0.7}{$\mathcal{N}(-5,0.5)$} & \scalebox{0.7}{$\mathcal{U}(-10,10)$} &  $2$  \\
		\rule{0pt}{3ex}$\mathbf{x}^{301} \cdots \mathbf{x}^{400}$ & \scalebox{0.7}{ $\mathcal{N}(5,0.5)$}  & \scalebox{0.7}{$\mathcal{U}(-10,10)$} & \scalebox{0.7}{$\mathcal{U}(-10,10)$} & \scalebox{0.7}{$\mathcal{N}(-5,0.5)$} & $2$  \\
		\hline
	\end{tabular}
\end{table}
Without   loss, we have assigned the first $200$ points to class $1$, and the last $200$ points to class 2. For class 1, instances $1$ to $100$ create a distinct cluster around $(0,-5)$ in the feature space formed of $x_{1},x_{2}$ and instances  $101$ to $200$ create a distinct cluster around $(0,0)$ in the feature space formed of $x_{3},x_{4}$.
Similarly, class 2 is also composed of two groups of points creating two distinct clusters in the feature spaces formed of $x_{2},x_{3}$ and $x_{1},x_{4}$ respectively.   To handle a dataset like Synthetic3 having within-class substructures we employ our proposed rule-specific approach implemented using (\ref{eq:mem_rule}), and (\ref{eq:sel_rs}).
\begin{table}[h!]
	\caption{Features subsets selected for synthetic3}
	\label{tab:tab7}
	\centering
	\begin{tabular}{|c|c|c|c|c|}
		\hline
		Run&\multicolumn{2}{c|}{Rule-specific} &\multicolumn{2}{c|}{Class-specific}\\\cline{2-5}
		& Class 1 & Class 2&  $\mathbf{s}_{1}$ &$\mathbf{s}_{2}$  \\
		\hline
		1 & rule 1:$x_{1},x_{2}$; rule 2:$x_{3},x_{4}$ & rule 1:$x_{2},x_{3}$;  rule 2:$x_{1},x_{4}$ & $x_{1}$ & $x_{1},x_{2}$ \\ 
		2 & rule 1:$x_{1},x_{2}$; rule 2:$x_{3},x_{4}$ & rule 1:$x_{1},x_{4}$;  rule 2:$x_{2},x_{3}$ & $x_{2}$ & $x_{1},x_{2}$ \\ 
		3 & rule 1:$x_{3},x_{4}$; rule 2:$x_{1},x_{2}$ & rule 1:$x_{1},x_{4}$;  rule 2:$x_{2},x_{3}$ & $x_{2}$ & $x_{2}$\\
		4 & rule 1:$x_{3},x_{4}$; rule 2:$x_{1},x_{2}$ & rule 1:$x_{1},x_{4}$;  rule 2:$x_{2},x_{3}$ & $x_{2}$ & $x_{1},x_{2}$\\
		5 & rule 1:$x_{3},x_{4}$; rule 2:$x_{1},x_{2}$ & rule 1:$x_{2},x_{3}$;  rule 2:$x_{1},x_{4}$ & $x_{2}$ & $x_{2}$\\
		\hline
	\end{tabular}
\end{table}
As observed from Table \ref{tab:tab7}, the rule-specific feature selection is successful in identifying the two important sub-spaces i.e. $x_{1},x_{2}$ and $x_{3},x_{4}$ for class 1 and $x_{2},x_{3}$ and $x_{1},x_{4}$ for class 2. It is noteworthy that for both the classes the selected rule-specific subsets interchange between rule 1 and 2. This is natural as the cluster number assignment to different groups of points for a class varies. We also note from Table \ref{tab:tab7}, using the class-specific feature selection method, in different runs, $\mathbf{s}_{1}$ comprises of $x_{1}$ or $x_{2}$ and $\mathbf{s}_{2}$ comprises of $x_{2}$ or $x_{1},x_{2}$. These subsets obviously do not characterize the classes correctly. The average accuracy of the FRBC with using feature subsets selected by rule-specific, class-specific feature selection and using all features are $100\%$, $77.4\%$, and $87.5\%$, respectively. 
This demonstrates  the usefulness of our proposed RSFS framework in data sets having multiple subspace-based structures or clusters within a class.  
\section{Conclusion}
In this work, first, we have proposed a class-specific feature selection scheme using  feature modulators embedded in a fuzzy rule-based classifier. The feature modulators can allow or stop the features from participating in the classification process by modifying their parameters. The feature modulator parameters are   tuned  by minimizing a loss function comprising of classification error and a regularizer to make the modulators completely select or reject features. This framework is used in \cite{chen2011integrated,chung2017feature} for selecting globally useful features. We modified it to make it suitable for CSFS. Our proposed class-specific feature selection method does not employ OVA strategy like most of the existing class-specific feature selection works and hence free from the enhanced computational overload and hazards associated with the existing OVA based methods. We have two more contributions. First, we have extended the CSFS scheme so that it can monitor class-specific redundancy by adding a suitable regularizer. Second, our CSFS framework is generalized to a rule-specific feature selection framework to handle the presence of multiple sub-space based structures or clusters within a class. All three approaches are validated through three experiments on appropriate synthetic data sets.    
\bibliographystyle{splncs04}
\bibliography{CSFS}
\end{document}